\documentclass[11pt]{article}

\usepackage[final]{acl}

  \usepackage{microtype}
  \usepackage{graphicx}
  \usepackage{amsmath}
  \usepackage{booktabs}
  \usepackage{multirow}
  \usepackage{tcolorbox}
\tcbuselibrary{skins,breakable}

\usepackage[english,bidi=default]{babel} 
\babelfont{rm}{TeXGyreTermesX} 
\babelprovide[import]{hindi}
\babelfont[*devanagari]{rm}{Lohit Devanagari}
\babelprovide[import]{arabic}
\babelfont[*arabic]{rm}{Noto Sans Arabic}

\babelprovide[import]{korean}
\babelprovide[import]{chinese}
\babelprovide[import]{japanese}

\babelfont[korean]{rm}[Path=fonts/Noto_Serif_KR/static/]{NotoSerifKR-Regular.ttf}
\babelfont[chinese]{rm}[Path=fonts/Noto_Serif_SC/static/]{NotoSerifSC-Regular.ttf}
\babelfont[japanese]{rm}[Path=fonts/Noto_Serif_JP/static/]{NotoSerifJP-Regular.ttf}

\title{Bridge Routing Heads: Where Multilingual Multi-hop\\Reasoning Lives in LLMs}

\author{
  \textbf{Seunghan Kim\textsuperscript{1}},
  \textbf{Minyeong Choe\textsuperscript{1}},
  \textbf{Hyunil Kim\textsuperscript{1}\thanks{Corresponding authors.}},
  \textbf{Haehyun Cho\textsuperscript{2}\footnotemark[1]}
\\
  \textsuperscript{1}Chosun University, 
  \textsuperscript{2}Soongsil University, 
\\
  \texttt{
  \textsuperscript{1}\{seunghan, minyeong, hyunil\}@chosun.ac.kr,
  \textsuperscript{2}haehyun@ssu.ac.kr
  }
}

\begin{document}
\maketitle
\begin{abstract}
Multilingual LLMs answer the same multi-hop reasoning question across languages, but we lack a mechanistic account of whether they share an internal circuit.
We identify \emph{Bridge Routing Heads} (BRH) in two large multilingual LLMs through a three-stage pipeline.
The resulting language-specific head sets exhibit near-complete mutual exclusivity across the five languages, with a mean Jaccard similarity of only 0.017 for Llama 3.1 70B and 0.057 for Qwen 2.5 72B, revealing language-idiosyncratic circuits.
Ablating general BRH increases two-hop Negative Log-Likelihood (NLL) by 39--89$\times$ the random-head baseline, providing direct causal evidence of their role.
Amplifying these heads in a failing target-language pass rescues up to 51.7\% of cross-lingual failures, with no training.
The two models share this dual-circuit pattern but allocate heads differently: Llama concentrates chaining in a large general pool, while Qwen leans on larger language-specific pools.
Together these results show that activation-level intervention alone can recover correct answers from cross-lingual reasoning failures.
\footnote{The code and data are available at \url{https://github.com/ICT-Convergence-Security-Lab-Chosun/bridge-routing-heads}}

\end{abstract}

\section{Introduction}
\label{sec:intro}

Recent advances in multilingual large language models (LLMs) enable them to handle queries across dozens of languages.
However, true multilingual capability requires more than simple translation: models must perform multi-step reasoning within each language.

A common test for this capability is the two-hop question, which requires retrieving an intermediate \emph{bridge entity} to derive the final answer (e.g., \emph{What is the currency of the country of [Person]?}).
Since these queries cannot be solved by simple pattern matching, they probe multi-step inference in LLMs~\cite{yang2018hotpotqa, trivedi2022musique, biran2024hopping}.

Motivated by this structure, a growing line of work has sought to understand the internal mechanisms underlying multi-hop reasoning~\cite{hou-etal-2023-towards, biran2024hopping, yu-etal-2025-back}.
More recently, these efforts have been extended to the multilingual setting, examining whether such reasoning mechanisms are shared or diverge across languages~\cite{ferrando2024similarity, zhang2024same, hu2025large}.
In contemporaneous work, \citet{sun2026bridging} attribute failures in lower-resource languages to misaligned attention pathways that fail to route information toward language-agnostic semantic neurons, and propose fine-tuning as a remedy.

However, existing studies are largely constrained to analyzing cross-lingual failures or basic grammatical processing.
The critical question of what the internal circuitry looks like when multilingual multi-hop reasoning actually succeeds in deriving the correct answer remains fundamentally unanswered.

In this work, we ask directly: \textit{when a multilingual LLM answers the same two-hop question correctly across languages, does it do so through the same internal circuit?} This distinction carries both theoretical and practical consequences.
A shared circuit would imply that the model has internalized a language-agnostic reasoning procedure; language-specific circuits would suggest that cross-lingual output consistency masks fundamentally divergent internal processes.
Our answer is neither: successful multilingual chaining is carried by a shared language-general head pool together with near-disjoint language-specific pools.

\begin{figure*}[t]
    \centering
    \includegraphics[width=\linewidth]{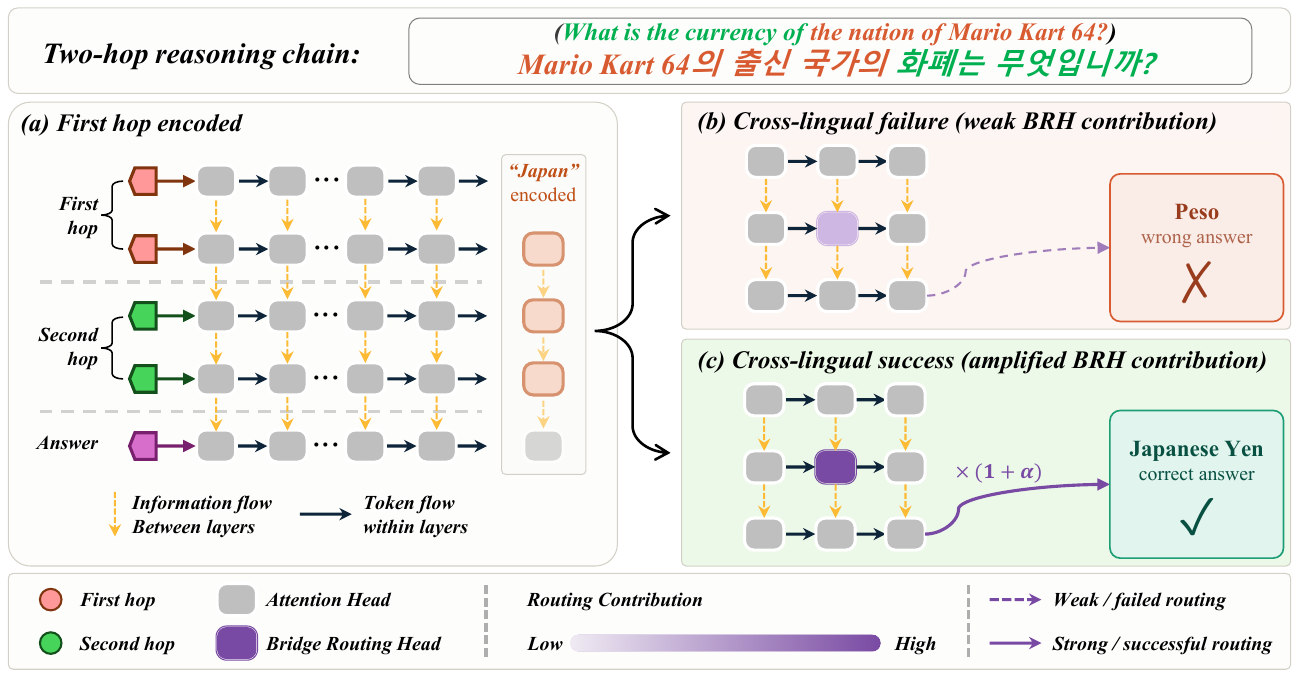}
    \caption{
        Overview of Bridge Routing Heads (BRH).
        \textbf{(a)} The first hop of a two-hop query resolves the bridge entity $B$ (here, \emph{Japan}) within the model's hidden states.
        \textbf{(b)} When the BRH routing contribution is under-weighted in the target language, the bridge signal is lost and the model produces an incorrect answer.
        \textbf{(c)} Amplifying BRH outputs by $1{+}\alpha$ restores routing of the bridge signal, recovering the correct answer---without any training.}
    \label{fig:overview}
\end{figure*}

We approach this question through mechanistic interpretability (MI), building on prior multilingual work on language-specific output neurons~\cite{tang2024language} and concept-space convergence~\cite{wendler2024llamas}.
We first localize the cross-lingual reasoning signal to the attention module (rather than MLP) via layer-wise activation patching, then ask whether the same attention heads drive successful multi-hop reasoning across languages.
We term these \textbf{Bridge Routing Heads (BRH)}---attention heads that carry an already-resolved bridge entity into the final answer---and test whether they causally drive the chaining process.
As illustrated in Figure~\ref{fig:overview}, whether BRH contribute strongly enough to the answer position determines the success or failure of cross-lingual multi-hop reasoning.

To answer this question, we construct a multilingual two-hop dataset from Wikidata following \citet{biran2024hopping}, covering five typologically diverse languages: English, Korean, Chinese, Japanese, and Spanish.
We then propose a three-stage pipeline---importance scoring, mean ablation, and Patchscopes verification---to identify BRH.
We validate the identified heads on two multilingual LLMs with contrasting pretraining language distributions: Llama 3.1 70B~\cite{grattafiori2024llama3herdmodels} and Qwen 2.5 72B~\cite{qwen2025qwen25technicalreport}.

\noindent
Our main contributions are as follows:
\begin{itemize}
    \item \textbf{BRH Identification.} We define Bridge Routing Heads (BRH) and identify them through a three-stage pipeline that separates a language-general head set from disjoint language-specific sets. 
    Language-specific head sets exhibit near-complete mutual exclusivity across five languages (mean Jaccard 0.017 for Llama; 0.057 for Qwen), revealing highly isolated, language-idiosyncratic circuits.
    \item \textbf{Causal Verification.} We provide causal evidence that ablating either BRH pool substantially raises two-hop NLL, and that amplifying them rescues failing target-language passes---intra-lingually through the language-specific pool, cross-lingually through the general pool---without any parameter update, demonstrating that such failures reflect an under-weighted routing circuit rather than missing knowledge.
    \item \textbf{Cross-model Analysis.} We show that both models share the same dual-circuit pattern---a language-general pool alongside disjoint language-specific pools---yet implement it differently: Llama 3.1 70B concentrates chaining in a larger general pool, while Qwen 2.5 72B relies on larger language-specific pools, a divergence that may reflect differences in their pretraining language mix, architecture, tokenization, post-training and others.
\end{itemize}

\section{Background \& Related Work}
\label{s:backg}

\noindent\textbf{MI module roles.}
Prior mechanistic interpretability work establishes that within each Transformer block, MLP layers act as key--value memories storing factual associations~\cite{geva-etal-2021-transformer,meng2022locating}, while attention heads route information between token positions~\cite{elhage2021mathematical}. 
In multilingual models, this division extends to language-specific phenomena: dedicated neurons govern output-language selection~\cite{tang2024language}, and representations converge toward a shared concept space across languages~\cite{wendler2024llamas,schut2025multilingual}, with non-Roman scripts further mediated by latent Romanization in mid-layers~\cite{saji2025romanlens}.
Whether the same module-level logic governs \emph{reasoning} circuitry, rather than surface linguistic behavior, remains an open question.

\noindent\textbf{Circuit discovery.}
Activation patching~\cite{wang2023interpretability} identifies circuits by measuring how component-level substitutions shift model behavior; ACDC~\cite{conmy2023towards} automates this via iterative edge pruning, requiring one forward pass per edge and becoming intractable at large scale.
We use activation patching to localize the responsible module type (\S\ref{s:stage1}), then switch to gradient-based importance scoring with a multi-condition contrast for head-level discovery (\S\ref{s:stage2}), which requires only one forward--backward pass per condition regardless of model size.

\noindent\textbf{Multi-hop reasoning circuits.}
\citet{biran2024hopping} dissected two-hop reasoning by showing that bridge entities are resolved in early layers while the second hop commences much later, identifying a ``hopping too late'' failure mode that motivates hop-separated head analysis.
The closest related work~\cite{sun2026bridging} identifies language-agnostic semantic neurons in cross-lingual multi-hop reasoning and attributes failures to misaligned attention pathways, proposing fine-tuning as a remedy.
We instead analyze the \emph{successful} multi-hop case across languages without parameter updates, isolating the attention circuits behind correct chaining rather than the failure mode behind incorrect ones.

\noindent\textbf{Multilingual multi-hop datasets.}
Dominant multi-hop QA benchmarks such as HotpotQA~\cite{yang2018hotpotqa} and MuSiQue~\cite{trivedi2022musique} ground each question in supporting paragraphs, conflating parametric retrieval with in-context reading and rendering them unsuitable for studying reasoning circuits.
Multilingual extensions compound the issue by translating English-sourced questions, introducing translation artifacts and label noise.
\section{Methodology}
\label{s:method}

Our pipeline answers the central question of \S\ref{sec:intro} in three stages on the multilingual two-hop dataset of \S\ref{s:data}: localize the module that carries cross-lingual reasoning (Stage 1), identify the attention heads through a three-condition importance contrast that yields \emph{Bridge Routing Heads} (Stage 2), and verify them causally with ablation and amplification (Stage 3).

\subsection{Multilingual Two-hop Dataset}
\label{s:data}

We build a multilingual two-hop reasoning dataset directly from Wikidata~\cite{10.1145/2629489}, without translation.
Each item composes two relations $A \xrightarrow{r_1} B \xrightarrow{r_2} C$ into a query that asks for $C$ given $A$, following the wiki-based construction of~\cite{biran2024hopping}.
We render the same composition in five languages chosen to vary along three typological axes---word order, family, and script (Table~\ref{tab:typology})---using each language's Wikidata labels.
The dataset contains 82{,}020 items indexed by Wikidata Q/P-IDs; full details are in Appendix~\ref{sec:data-structure}.

\begin{table}[t]
  \centering
  \footnotesize
  \setlength{\tabcolsep}{4pt}
  \begin{tabular}{l c l l}
    \hline
    Language & Order & Family & Script \\
    \hline
    English (EN)  & SVO & IE (Germanic) & Latin \\
    Korean (KO)   & SOV & Koreanic      & Hangul \\
    Chinese (ZH)  & SVO & Sino-Tibetan  & Han \\
    Japanese (JA) & SOV & Japonic       & Kana + Han \\
    Spanish (ES)  & SVO & IE (Romance)  & Latin \\
    \hline
  \end{tabular}
  \caption{Typological coverage of the five languages in our dataset. IE = Indo-European; Han = Han / Chinese characters.}
  \label{tab:typology}
\end{table}

\noindent\textbf{Prompts and controls.}
Each composition yields five prompts: the two-hop query, two single-hop queries (positive controls, each isolating one relation), and two shortcut queries that drop either $A$ or the first relation (negative controls---a model relying on genuine multi-hop composition should fail these; detailed examples in Appendix~\ref{sec:data-structure}, Table~\ref{tab:query_types}).

\noindent\textbf{Knowledge-verified pool.}
To study routing failures rather than knowledge gaps, we restrict every analysis to items whose required facts the model already holds.
For each model and language $X$, we retain an item only when both single-hop queries are answered correctly in $X$ and both shortcut queries are answered incorrectly in all five languages; we call the surviving items the \emph{knowledge-verified pool} $\mathcal{K}_X$ (sizes in Table~\ref{tab:setup-counts}).

\noindent\textbf{Entity policy and verification.}
While localizing $A$ into each target language does not substantially change our results (Appendix~\ref{app:entity-localization}), we keep the source entity $A$ in its English surface form across all languages (e.g., ``Stephen Harper'' appears in the Korean, Chinese, Japanese, and Spanish prompts), isolating reasoning from per-language proper-noun knowledge gaps and keeping the entity token position consistent.
We accept an item only when the gold answer matches the same Wikidata entity in all five languages, so cross-lingual differences arise from internal computation, not data drift.

\subsection{Layer-wise Activation Patching}
\label{s:stage1}

The first stage isolates which module type---MLP or attention---carries the cross-lingual reasoning signal.
We feed each two-hop question in a direct QA format (\texttt{Instruction: Answer briefly with only the answer. Question: \{q\}? Answer:}), forcing the model to rely on parametric knowledge rather than in-context retrieval; the instruction prefix is necessary to suppress multiple-choice format drift that occurs in its absence.

\noindent\textbf{Patching.}
For each language pair and layer $\ell$, we cache the source-language residual stream at layer $\ell$ and position \texttt{answer\_colon} (the final \texttt{:} token of the prompt template).
We then substitute the target-language residual stream at the same layer and position with the cached value.
We perform this substitution separately for MLP blocks and attention blocks, following the established view that MLP blocks store factual associations~\cite{geva-etal-2021-transformer,meng2022locating} and attention heads move information between positions~\cite{elhage2021mathematical}, and sweep all layers without a prior hypothesis.

\noindent\textbf{Metrics.}
We score each patch with KL divergence $D_{\mathrm{KL}}(P_{\mathrm{clean}} \,\|\, Q_{\mathrm{patched}})$ (magnitude of output-distribution change) and logit difference $\mathrm{LD} = \mathrm{logit}_{\mathrm{patched}}[t^\star] - \mathrm{logit}_{\mathrm{clean}}[t^\star]$ (directional effect on the clean top-1 token $t^\star$; positive strengthens, negative suppresses).
For each module type, we compute the Pearson correlation between per-layer KL and LD profiles across language pairs to examine how consistently each module type responds across languages.

\subsection{Bridge Routing Head Detection}
\label{s:stage2}

The second stage identifies which attention heads carry the resolved bridge entity into the final-answer prediction.
We adopt a three-condition differential design: a genuine chaining head should contribute when the bridge must be inferred (first-hop and two-hop conditions) but not when it is supplied externally (second-hop condition), isolating chaining from whole-task importance.
Candidates pass through three sequential filters.

\noindent\textbf{Step 1: Importance scoring.}
LAHIS~\cite{Liu_Song_Zhou_Du_Xu_Jiang_Zhang_Jia_2026} estimates per-head importance via the gradient of the task loss with respect to a soft mask on each head's output projection, requiring one forward and one backward pass per condition.
For each head $h$, we collect importance scores under three conditions: \textbf{FH} (first-hop, asks for bridge $B$ given $A$ and $r_1$), \textbf{TH} (two-hop, asks for $C$ given $A$ and both relations without revealing $B$), and \textbf{SH} (second-hop, asks for $C$ given $B$ and $r_2$ directly).
We combine these into the Bridge Routing Score (BRS):
\[
  \mathrm{BRS}(h) = z(I_{\mathrm{FH}}) + z(I_{\mathrm{TH}}) - z(I_{\mathrm{SH}}),
\]
where $z(\cdot)$ z-normalizes importance over all heads within each condition, removing magnitude differences across conditions.
The top $N$\% of heads by BRS---with $N$ tuned per model so the strict cross-language intersection covers 3--4\% of the head budget---form per-language pools; their strict intersection across all five languages is the \emph{general candidate pool}.
Masking the general pool and taking the top 4\% of remaining heads per language yields the \emph{language-specific candidate pool}.
We validate this contrast in Appendix~\ref{app:brs-contrast}: the TH term alone dominates the per-head ranking (Spearman $\rho = 0.962$ against the original), while the FH$+$ and $-$SH terms narrow the candidate pool to its chaining-specific subset.

\noindent\textbf{Step 2: Mean-ablation filter.}
We ablate each candidate head individually by replacing its output with a pre-computed mean activation---the average head output over a reference set of 450 correct two-hop prompts (30 items per condition per language, across all three conditions and five languages).
Following \citet{wang2023interpretability}, we use mean ablation rather than zero ablation, as zeroing produces activations outside the model's normal distribution and can trigger compensatory re-routing by backup mechanisms.
We then record $\Delta\mathrm{NLL} = \mathrm{NLL}_{\mathrm{ablated}} - \mathrm{NLL}_{\mathrm{clean}}$, where NLL is the negative log-likelihood of the gold answer token; $\Delta\mathrm{NLL} > 0$ therefore indicates that the head was causally contributing to the correct prediction.
Candidates with $\Delta\mathrm{NLL} \leq 0$ are dropped.

\noindent\textbf{Step 3: Patchscopes verification.}
For the language-specific candidates that survive Step~2, we verify whether the hidden state of the layer containing each head encodes the bridge entity $B$ using Patchscopes~\cite{patchscopes2024}. 
As illustrated in Figure~\ref{fig:patchscopes-example}, this layer-level hidden state is injected into a few-shot interpretation prompt at the position labeled \texttt{x:}. A candidate head passes this verification if the bridge entity $B$ successfully emerges anywhere in the subsequent free generation.
Surface leakage from the in-context examples can occur as a benign artifact, so a single success across trials suffices.
We call heads that survive all applicable filters \emph{Bridge Routing Heads} (BRH): general heads pass Steps 1--2, language-specific heads pass all three.

\begin{figure}[t]
  \centering
  \begin{tcolorbox}[
    width=\linewidth,
    colback=gray!4, colframe=black!55,
    boxrule=0.5pt, arc=2pt,
    left=6pt, right=6pt, top=4pt, bottom=4pt,
    title={\footnotesize\bfseries Patchscopes example: Llama 3.1 70B, Korean-specific BRH},
    fonttitle=\footnotesize\bfseries,
    coltitle=white, colbacktitle=black!55,
  ]
  \footnotesize
  \noindent\textbf{Two-hop question (KO):} \foreignlanguage{korean}{Rod Laver의 국적의 수도는 무엇입니까?}\par
  \emph{Capital of Rod Laver's nationality?} \\
  \textbf{Chain:} \\
  Rod Laver $\xrightarrow{\text{nat.}}$ \textbf{Australia} $\xrightarrow{\text{cap.}}$ Canberra. \\
  Gold bridge $B$ = Australia.

  \medskip
  \noindent\textbf{Interpretation prompt} (BRH hidden state patched at \texttt{x:}):\par
  \texttt{Syria: ...; Leonardo DiCaprio: ...; Samsung: ...; x:}

  \medskip
  \noindent\textbf{Model output:} \emph{``Australia: Australia is a country in the southern hemisphere, ...''}
  \end{tcolorbox}
  \caption{One Patchscopes example for a Llama 3.1 70B Korean-specific BRH. The patched hidden state at \texttt{x:} causes the model to generate the gold bridge ``Australia'' despite the question being in Korean---direct evidence that this BRH encodes the bridge entity. Full set: Table~\ref{tab:patchscopes}.}
  \label{fig:patchscopes-example}
\end{figure}

\subsection{Verifying BRH}
\label{s:stage3}

The third stage asks whether BRH actually drive reasoning, through a head-set overlap analysis, ablation tests, and amplification interventions.

\noindent\textbf{Head-set overlap.}
We compute Jaccard similarity for all pairs of language head sets per model to characterize how much the language-specific sets share with one another.
A near-zero Jaccard similarity between any two language-specific sets would confirm that each language's pool targets mechanistically distinct components, rather than merely reflecting a common importance-ranking artifact.

\noindent\textbf{Ablation.}
Following the same mean-ablation protocol as Step~2 (\S\ref{s:stage2}), we neutralize each head's output using a mean computed from 50 correct two-hop prompts per language (250 per model total)---two-hop items only, since the verification target is whole-chain performance rather than individual-hop sensitivity.
Rather than filtering individual heads, we ablate the general BRH pool and each language-specific pool as a unit on 100 correct two-hop items per language, recording the aggregate $\Delta\mathrm{NLL}$.
We compare against a layer-matched random head set of the same size (20 repeats); a larger $\Delta\mathrm{NLL}$ for BRH confirms collective causal responsibility for routing the bridge entity.

\noindent\textbf{Amplification.}
We scale BRH outputs by $1 + \alpha$ ($\alpha \in \{0.25, 0.5, 1.0\}$) during the target-language forward pass on items the model originally answered incorrectly, and measure the rescue rate as a function of $\alpha$.
As in the ablation, we compare each amplification against layer-matched random head sets of the same size (20 repeats).
\begin{itemize}
  \item \emph{Intra-lingual Amplification}: For each language $X$, we amplify the language-specific BRH on all items the model answered incorrectly in $X$. A monotone increase in accuracy with $\alpha$ provides direct causal evidence that the magnitude of the language-specific pool drives target-language performance.
  \item \emph{Cross-lingual Transfer}: We amplify only the general BRH on EN-success $\cap$ $X$-fail items---items the model answers correctly in English but fails in $X$. Since the model produced the correct answer in English, it demonstrably holds the relevant factual knowledge; any $X$-language rescue therefore reflects an under-weighted routing circuit rather than missing knowledge, and the baseline accuracy is $0$ by construction.
\end{itemize}
\section{Experiments \& Results}
\label{s:eval}

\subsection{Setup}

\noindent\textbf{Models.}
We apply this pipeline to two instruction-tuned 70B-class models with contrasting pretraining language mixes---Llama 3.1 70B (English-centric) and Qwen 2.5 72B (Chinese-strong)---to test whether the dual-circuit pattern holds across pretraining regimes; English serves as the cross-lingual pivot.

\begin{table}[t]
  \centering
  \scriptsize
  \setlength{\tabcolsep}{1.5pt}
  \begin{tabular}{l l r r r r r}
    \hline
    Model & Pool & EN & KO & ZH & JA & ES \\
    \hline
    Llama & Knowledge-verified ($\mathcal{K}_X$) & 10{,}059 & 7{,}106 & 6{,}672 & 6{,}982 & 7{,}967 \\
          & Two-hop correct                       & 3{,}538  & 2{,}467 & 1{,}981 & 2{,}290 & 2{,}653 \\
          & Two-hop incorrect                     & 6{,}521  & 4{,}639 & 4{,}691 & 4{,}692 & 5{,}314 \\
          & EN correct / $X$ incorrect             & ---      & 519     & 489     & 420     & 582 \\
    \hline
    Qwen  & Knowledge-verified ($\mathcal{K}_X$) & 4{,}101 & 7{,}395 & 2{,}654 & 5{,}994 & 8{,}960 \\
          & Two-hop correct                       & 1{,}700 & 1{,}977 & 795     & 1{,}952 & 2{,}242 \\
          & Two-hop incorrect                     & 2{,}401 & 5{,}418 & 1{,}859 & 4{,}042 & 6{,}718 \\
          & EN correct / $X$ incorrect             & ---     & 244     & 79      & 209     & 472 \\
    \hline
  \end{tabular}
  \caption{Per-language sizes of the knowledge-verified pool and experiment-specific subsets.}
  \label{tab:setup-counts}
\end{table}

\noindent\textbf{Data subsets.}
Following the knowledge verification in \S\ref{s:data}, every experiment draws from $\mathcal{K}_X$ (Table~\ref{tab:setup-counts}).
BRH scoring, filtering, Patchscopes, and ablation use items answered correctly on the target-language two-hop query, whereas Intra-Lingual Amplification uses items answered incorrectly.
Cross-Lingual Transfer further restricts those failures to items answered correctly in English.

\begin{table*}[t]
  \centering
  \footnotesize
  \begin{tabular}{l l r r r r}
    \hline
    Model & Set & After scoring & After ablation & After Patchscopes & Final (\% of total) \\
    \hline
    Llama 3.1 70B & general & 191 & 70  & ---   & 70 (1.37\%) \\
    (5,120 Heads) & ko-spec & 204 & 114 & 8     & 8 (0.16\%) \\
                  & zh-spec & 204 & 115 & 11    & 11 (0.21\%) \\
                  & ja-spec & 204 & 114 & 31    & 31 (0.61\%) \\
                  & es-spec & 204 & 120 & 9     & 9 (0.18\%) \\
    \hline
    Qwen 2.5 72B  & general & 175 & 42  & ---   & 42 (0.82\%) \\
    (5,120 Heads) & ko-spec & 204 & 122 & 87    & 87 (1.70\%) \\
                  & zh-spec & 204 & 93  & 38    & 38 (0.74\%) \\
                  & ja-spec & 204 & 111 & 28    & 28 (0.55\%) \\
                  & es-spec & 204 & 107 & 31    & 31 (0.61\%) \\
    \hline
  \end{tabular}
  \caption{Head counts at each stage of the BRH filtering pipeline. After scoring uses the top-$N$\% threshold from Step~1 (\S\ref{s:stage2}): top 10\% for Llama, top 13\% for Qwen.}
  \label{tab:brh-counts}
\end{table*}

\noindent\textbf{Statistics.}
We report 95\% Wilson CIs for accuracy, 95\% bootstrap CIs for $\Delta\mathrm{NLL}$ (1,000 resamples), and compare ablations against 20 layer-matched random head sets.

\subsection{Module-level Localization}
\label{s:exp-stage1}

\begin{figure}[!t]
  \centering
  \includegraphics[width=\linewidth]{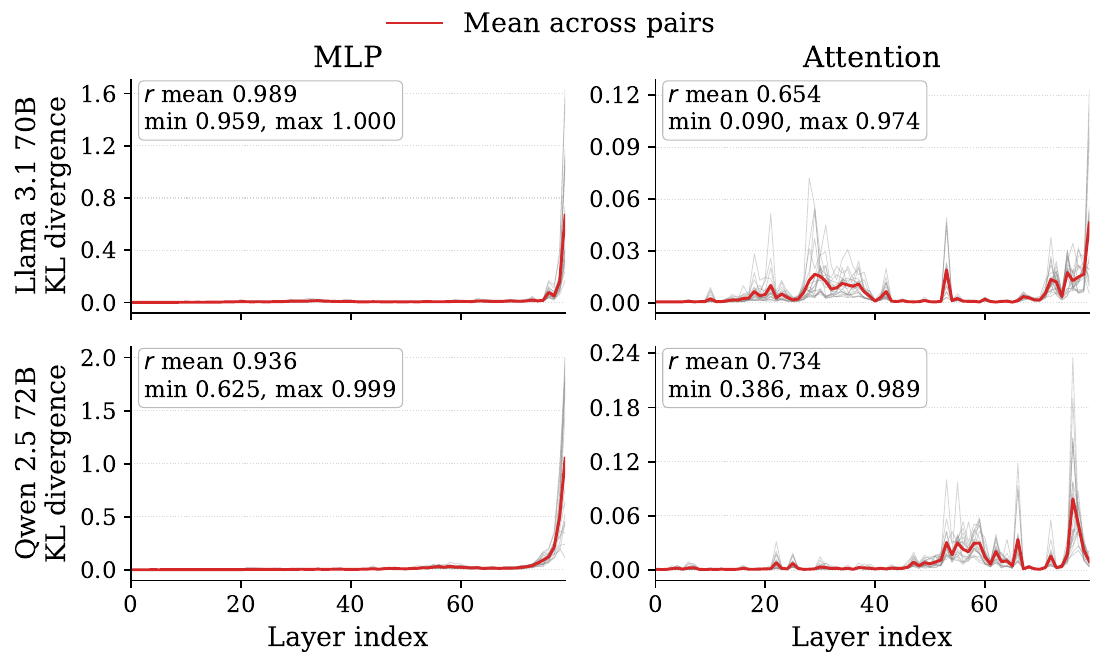}
  \caption{Layer-wise activation patching on the five-language dataset, KL divergence at \texttt{answer\_colon}. $r$ means the mean Pearson correlation coefficient with range (min, max) over the 380 off-diagonal pairs. Each panel overlays the 20 directional language-pair curves (grey) and their mean (red).}
  \label{fig:stage1-sweep}
\end{figure}

\noindent\textbf{MLP curves are nearly identical across language pairs.}
KL profiles of the MLP patch are tightly coupled across the 20 directional pairs in both models (Figure~\ref{fig:stage1-sweep}, left column).
The shared shape---near-zero KL in early layers, a steep climb after layer $\sim$60, and a sharp final-layer spike---persists regardless of which language pair we patch, consistent with the view that MLP blocks store language-agnostic factual associations.

\noindent\textbf{Attention curves diverge.}
The same correlations on attention drop in both models (Figure~\ref{fig:stage1-sweep}, right column).
The fan-out of individual pair curves is visible across the entire mid-to-late layer range, indicating that attention carries language-pair-specific routing behavior that MLP does not.
The heterogeneous attention signal motivates the head-level analysis in Stage 2.
Logit-difference profiles for the same experiment, which corroborate these conclusions, are provided in Appendix~\ref{sec:layer-wise-patching}.

\subsection{BRH Identification}
\label{s:exp-stage2}

\begin{table*}[t]
  \centering
  \footnotesize
  \setlength{\tabcolsep}{3pt}
  \begin{tabular}{l l r r r r r r r r}
    \hline
    \noalign{\vspace{2pt}}
    & & & \multicolumn{3}{c}{\textbf{General BRH}} & & \multicolumn{3}{c}{\textbf{Language-Specific BRH}} \\
    \cline{4-6}\cline{8-10}
    \noalign{\vspace{2pt}}
    Model & Lang & Base NLL & $\Delta$NLL (95\% CI) & Rand $\Delta$ & Scale &  & $\Delta$NLL (95\% CI) & Rand $\Delta$ & Scale \\
    \hline
    Llama 3.1 70B & ko & 0.72 & \textbf{2.07 (1.85--2.30)} & 0.053 & 39$\times$ & &\textbf{0.12 (0.09--0.15)} & 0.008 & 16$\times$ \\
                  & zh & 0.78 & \textbf{1.98 (1.73--2.23)} & 0.048 & 41$\times$ & &\textbf{0.08 (0.05--0.10)} & 0.010 & 8$\times$ \\
                  & ja & 0.70 & \textbf{1.68 (1.51--1.87)} & 0.040 & 42$\times$ & &\textbf{0.33 (0.28--0.38)} & 0.031 & 11$\times$ \\
                  & es & 0.85 & \textbf{1.87 (1.66--2.10)} & 0.042 & 45$\times$ & &\textbf{0.13 (0.11--0.16)} & 0.011 & 12$\times$ \\
    \hline
    Qwen 2.5 72B  & ko & 0.95 & \textbf{1.15 (1.02--1.31)} & 0.014 & 83$\times$ & &\textbf{0.65 (0.58--0.73)} & 0.017 & 37$\times$ \\
                  & zh & 1.11 & \textbf{1.02 (0.82--1.23)} & 0.020 & 52$\times$ & &\textbf{0.93 (0.77--1.08)} & 0.012 & \textbf{74$\times$} \\
                  & ja & 0.64 & \textbf{0.92 (0.83--1.02)} & 0.012 & 75$\times$ & &\textbf{0.50 (0.44--0.56)} & 0.021 & 25$\times$ \\
                  & es & 1.04 & \textbf{0.84 (0.71--0.99)} & 0.009 & \textbf{89$\times$} & &\textbf{0.54 (0.46--0.63)} & $-$0.005 & --- \\
    \hline
  \end{tabular}
  \caption{Mean-ablation results with 95\% bootstrap CIs. The Scale column reports how many times larger the BRH effect is than the random baseline; dashes indicate cases where the random baseline is non-positive, making the ratio undefined.}
  \label{tab:ablation}
\end{table*}

\noindent\textbf{General and specific heads occupy disjoint sets.}
We compute the Jaccard similarity for all 10 pairs of head pools per model (general, ko, zh, ja, es).
The general pool shares zero heads with every language-specific pool in both models, by construction (\S\ref{s:stage2}).
The remaining language-language pairs are sparse.
Llama has a mean Jaccard of 0.017, with non-zero overlap only in ko-ja (0.054) and zh-ja (0.050).
Qwen has a higher mean of 0.057, with non-zero overlap in all six pairs and a peak at ko-zh (0.097).
The qualitative pattern follows the typological grouping: Qwen, whose pretraining is heavier on Chinese, shares more heads across the CJK languages than across the SVO Latin-script ones.

\noindent\textbf{Final head counts differ sharply across models.}
As shown in Table~\ref{tab:brh-counts}, Llama has a larger general pool but smaller language-specific pools; Qwen has the inverse profile.
The most striking case is Korean: Qwen retains 87 specific BRH for Korean against Llama's 8, an order of magnitude apart.

\begin{figure}[t]
  \centering
  \includegraphics[width=\linewidth]{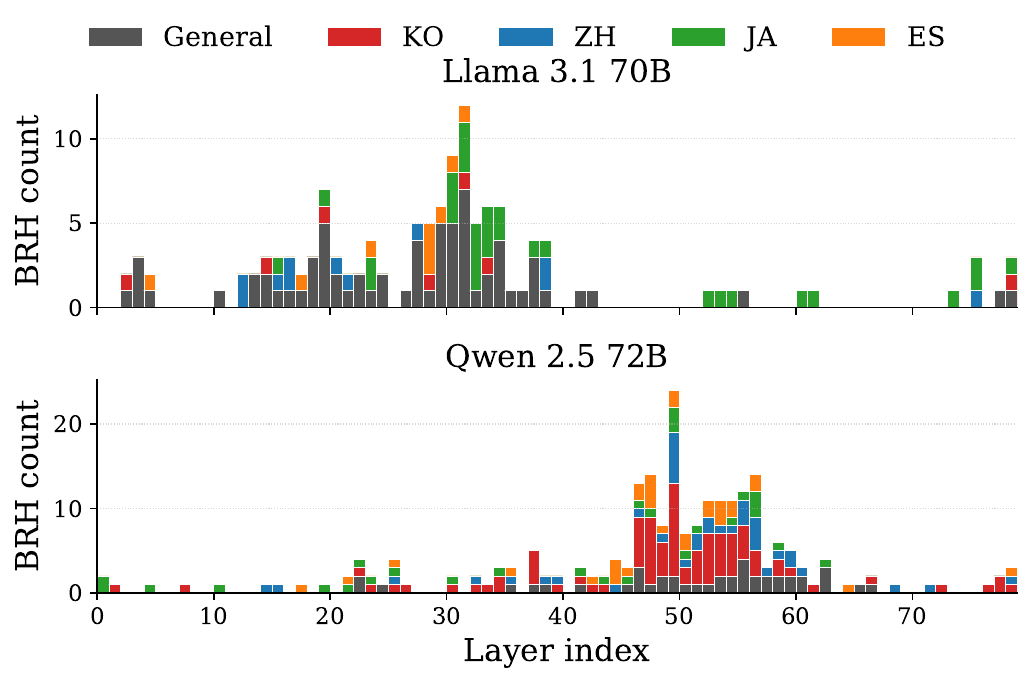}
  \caption{Layer position of every final BRH, stacked by set type.}
  \label{fig:brh-layers}
\end{figure}

\noindent\textbf{BRH concentrate in narrow layer bands.}
Figure~\ref{fig:brh-layers} plots the layer position of every final BRH for both models.
In Llama, BRH cluster sharply around layers 14--38, with general heads dominating the central peak and a sparse tail of Japanese-specific heads extending into late layers.
In Qwen, the distribution shifts later: layers 47--57 form a pronounced peak dominated by Korean-specific heads, with the general pool likewise concentrated in this region and smaller contributions from the remaining language-specific pools.
In both models, the BRH concentration bands align with the layers where their respective attention patching curves diverged most in Figure~\ref{fig:stage1-sweep}, suggesting the two analyses converge on the same routing region.
The same bands are visible in the unthresholded per-head BRS landscape, before any filtering step (Appendix~\ref{sec:bridge-routing-heads-heatmap}).

\subsection{Causal Verification}
\label{s:exp-stage3}

\begin{table*}[t]
  \centering
  \scriptsize
  \setlength{\tabcolsep}{4pt}
  \begin{tabular}{l l r r r r r r}
    \hline
    & & \multicolumn{3}{c}{\textbf{Intra-Lingual Amplification}} & \multicolumn{3}{c}{\textbf{Cross-Lingual Transfer}} \\
    Model & Lang & $\alpha{=}0.25$ & $\alpha{=}0.5$ & $\alpha{=}1.0$ & $\alpha{=}0.25$ & $\alpha{=}0.5$ & $\alpha{=}1.0$ \\
    \hline
    Llama 3.1 70B & ko & 2.0 (1.6--2.4) & 3.5 (3.0--4.1) & 6.1 (5.5--6.8) & 22.2 (18.8--25.9) & 30.4 (26.6--34.5) & 41.0 (36.9--45.3) \\
                  & zh & 2.6 (2.1--3.1) & 4.8 (4.2--5.5) & 8.6 (7.9--9.5) & 25.0 (21.3--29.0) & 35.6 (31.5--39.9) & 51.5 (47.1--55.9) \\
                  & ja & 3.5 (3.0--4.0) & 5.9 (5.3--6.6) & 10.1 (9.3--11.0) & 22.9 (19.1--27.1) & 37.9 (33.4--42.6) & \textbf{51.7 (46.9--56.4)} \\
                  & es & 1.4 (1.1--1.8) & 2.9 (2.5--3.4) & 4.8 (4.3--5.5) & 29.2 (25.7--33.0) & 36.8 (33.0--40.8) & 36.9 (33.1--40.9) \\
    \hline
    Qwen 2.5 72B  & ko & 7.8 (7.1--8.5) & 12.3 (11.4--13.2) & \textbf{17.2 (16.3--18.3)} & 15.6 (11.6--20.7) & 20.1 (15.5--25.6) & 23.0 (18.1--28.6) \\
                  & zh & 3.6 (2.9--4.6) & 5.5 (4.5--6.6) & 8.1 (7.0--9.5) & 17.7 (10.9--27.6) & 24.1 (16.0--34.5) & 31.7 (22.5--42.6) \\
                  & ja & 3.5 (3.0--4.1) & 5.7 (5.0--6.4) & 7.9 (7.1--8.8) & 5.7 (3.3--9.8) & 8.1 (5.1--12.6) & 12.0 (8.2--17.1) \\
                  & es & 6.8 (6.2--7.4) & 10.9 (10.2--11.7) & 14.7 (13.9--15.6) & 26.9 (23.1--31.1) & 34.5 (30.4--38.9) & 34.8 (30.6--39.2) \\
    \hline
  \end{tabular}
  \caption{Amplification accuracies (\%) with 95\% Wilson CIs. Intra-Lingual Amplification uses target-language failures; Cross-Lingual Transfer further requires a correct English answer. Subset sizes are given in Table~\ref{tab:setup-counts}.}
  \label{tab:amplification}
\end{table*}

\noindent\textbf{General BRH carry the chaining signal.}
Mean-ablating each model's general BRH increases two-hop NLL by 0.84--2.07 above the unpatched baseline (Table~\ref{tab:ablation}).
The same intervention with a layer-matched random head set produces a $\Delta\mathrm{NLL}$ of at most 0.053---the general BRH effect exceeds the random baseline by 39--89$\times$.
The result rules out chance and pins the chaining function to the general pool.

\noindent\textbf{Language-specific BRH show smaller but consistent effects.}
Mean-ablating the language-specific pool raises NLL above the random baseline in every (model, language) cell (Table~\ref{tab:ablation}).
Qwen's specific pools yield the larger effects ($\Delta\mathrm{NLL}$ 0.50--0.93); Llama's specific pools are smaller in size (8--31 heads vs.\ Qwen's 28--87) and their effects are correspondingly modest (0.08--0.33).
While these magnitudes are well below the general-pool effects, every cell remains positive and exceeds the random baseline by at least 8$\times$, indicating that the language-specific pool also makes a measurable causal contribution---consistent with the dual-circuit picture, with the magnitude of each pool's contribution tracking its head count.

\noindent\textbf{Patchscopes verifies bridge-entity encoding.}
Patchscopes further confirms that the surviving language-specific heads encode the gold bridge entity $B$ in their hidden states (Figure~\ref{fig:patchscopes-example}; full outputs in Table~\ref{tab:patchscopes} in Appendix~\ref{sec:appendix-patchscopes}).
Tightening the criterion from $B$ recovered in at least one trial to at least three of five trials shrinks both models' language-specific pools but preserves the Qwen $>$ Llama ordering at every threshold (Appendix~\ref{app:patchscopes-threshold}).

\begin{figure}[t]
  \centering
  \includegraphics[width=\linewidth]{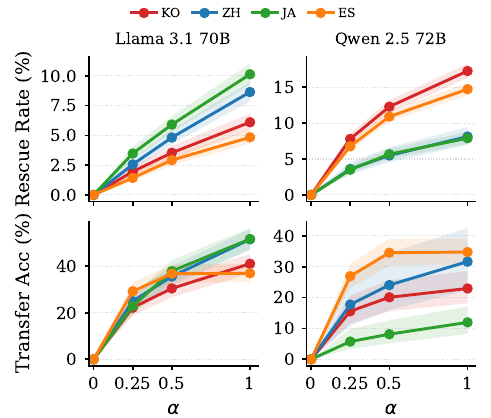}
  \caption{Accuracy as a function of the amplification scale $\alpha$. \emph{Top}: Intra-lingual amplification (rescue rate). \emph{Bottom}: Cross-lingual transfer.}
  \label{fig:scaling-transfer}
\end{figure}

\noindent\textbf{Amplifying specific BRH rescues incorrect items, training-free.}
Amplifying each language's specific BRH output by $1 + \alpha$ on target-language failures raises accuracy monotonically in $\alpha$ across every cell (Table~\ref{tab:amplification}; top row of Fig.~\ref{fig:scaling-transfer}).
At $\alpha = 1.0$, the rescue rate over the incorrect pool runs from 4.8\% (Llama-es) to 17.2\% (Qwen-ko), corresponding to a 9.7--20.7\% increase over Llama's original correct pool and 16.3--47.2\% over Qwen's---all with no parameter update.
Layer-matched random head sets of the same size rescue only 1.8--2.8\% (Llama) and 1.6--4.0\% (Qwen), so the specific-BRH effect exceeds the random baseline by 2.1--4.9$\times$.
The monotonic $\alpha$ dependence is direct causal evidence that specific BRH magnitude drives target-language accuracy.

\noindent\textbf{General BRH transfer reasoning across languages.}
Amplifying only the general BRH (Table~\ref{tab:amplification}; bottom row of Fig.~\ref{fig:scaling-transfer}) recovers 41--51.7\% of failures in Llama (mean 45.3\%) and 12--34.8\% in Qwen (mean 25.3\%), peaking at 51.7\% on Llama-ja.
The same intervention with layer-matched random head sets transfers only 10.2\% (Llama) and 5.4\% (Qwen)---the general BRH means exceed the random baseline by 4.4--4.7$\times$.
Unlike the CJK languages, whose performance continues to improve beyond $\alpha = 0.5$, Spanish largely saturates at this point in both models, showing only marginal gains thereafter.
The model is not missing knowledge in these cases---it has the answer accessible through the general BRH, and we are turning their contribution up.
These results indicate that some routing failures are correctable by activation amplification alone, without any retraining.
General-domain ability is largely preserved under the same intervention, with English MMLU~\cite{hendrycks2021measuring} accuracy changing by at most 2.1 points across all amplified head sets (Appendix~\ref{app:amplification-cost}).

\subsection{Model Comparison}
\label{s:exp-stage4}
The two models share the dual-circuit pattern but allocate heads differently.
Llama uses a large general pool (70 heads) with thin language-specific pools (8--31); Qwen has the inverse profile (general 42, specific 28--87).
The interventions mirror this allocation: Llama's general BRH transfer reasoning more effectively (mean 45.3\% vs.\ Qwen's 25.3\%), while Qwen's specific BRH amplification recovers more incorrect items (mean 11.9\% vs.\ Llama's 6.1\%).
\section{Discussion}
\label{s:discussion}

\noindent\textbf{Training-free intervention.}
Our amplification result shows the relevant attention circuits can be re-aligned at inference time, with no parameter update.
Prior work diagnoses misalignment at bridge resolution but proposes targeted fine-tuning~\cite{sun2026bridging}; the practical question shifts from teaching the model to surfacing what it already knows.

\noindent\textbf{Attention-side counterpart of the concept space.}
BRH may be the attention-side counterpart of the multilingual concept space~\cite{wendler2024llamas,schut2025multilingual}.
The general BRH route bridge entities through this shared, language-agnostic workspace; the language-specific BRH handle per-language recovery into the target output.

\noindent\textbf{Why amplification works.}
Amplification rescues failures because the general BRH already carry sufficient information for cross-lingual chaining---their contribution is merely under-weighted in the failing pass.
This under-use is distinct from post-ablation backup-head compensation~\cite{wang2023interpretability}.
General-BRH output norms are near-identical on success and failure passes (within $\sim$1\%), so the failing pass does not carry a weaker signal; because the norm does not constrain routing direction, alternative explanations (e.g., routing misalignment) remain compatible.
Pinpointing the failure mechanism will require direct success--failure comparisons of BRH output directions together with in-context bridge readouts on failing passes, which we leave to future work.

\noindent\textbf{Open questions.}
Two threads our experiments do not close.
First, head allocation across models: Llama and Qwen differ in pretraining language mix, architecture, tokenization, and post-training, so the observed allocation contrast cannot be attributed to pretraining mix alone---controlled comparisons across additional model families are needed to disentangle these factors.
Second, cross-language asymmetries: Spanish appears to saturate by $\alpha = 0.5$ in both models, whereas Qwen's transfer rates at $\alpha = 1.0$ vary substantially across the CJK languages (12--32\%).
These patterns could reflect lexical and script overlap with English, similarities in word order and other typological properties, or differences in pretraining exposure; our analysis does not distinguish among these factors, leaving typological proximity as a hypothesis rather than an established explanation.

\section{Conclusion}
\label{s:conclusion}

We asked whether multilingual LLMs share an internal circuit when answering the same multi-hop question correctly across languages.
Our answer is neither fully shared nor fully language-specific.
We defined \emph{Bridge Routing Heads} (BRH) in Llama 3.1 70B and Qwen 2.5 72B through a three-stage pipeline (hop-differential scoring, mean ablation, Patchscopes verification), where they split into a language-general pool and disjoint language-specific pools.
Despite differing allocations between the general and language-specific pools, the interventions yield strong causal effects in both models: ablating the general pool raises two-hop NLL by 39--89$\times$ the random-head baseline, and amplifying it rescues up to 51.7\% of cross-lingual failures without training.
These results show that some target-language failures reflect an under-weighted routing circuit rather than missing knowledge, opening a path to training-free multilingual model editing.

\section*{Limitations}
\label{s:limitations}

\noindent\textbf{Set-level interventions.}
Our causal claims rest on set-level mean ablation, which averages over heads within a pool and can hide heterogeneity between them.
Individual heads may contribute to chaining at different magnitudes, or even act in opposite directions, particularly in the smaller language-specific pools where a few heads dominate the aggregate.
Per-head causal decomposition is a natural extension of the present work.

\noindent\textbf{Two-model evaluation.}
Our cross-model evidence rests on two models.
Llama 3.1 70B and Qwen 2.5 72B differ along several axes at once---pretraining language mix, architecture, tokenization, and post-training---so fully testing generality across multilingual LLMs requires additional model families.
The natural next axis is models with other dominant languages (e.g., Korean- or Japanese-strong instruction-tuned LLMs).

\noindent\textbf{Language coverage.}
We cover five languages chosen for typological diversity along three axes (word order, family, script).
Low-resource languages with very different morphological systems, or languages without sufficient Wikidata coverage, are out of scope.

\noindent\textbf{Patchscopes verification threshold.}
The Patchscopes check we use is a verification step on heads that already pass the mean-ablation stage, not a strict filter.
A head qualifies once it generates the gold bridge entity at least once across the trials per item, which means heads that encode the bridge only intermittently can pass.
We argue the permissive criterion is appropriate here, because Patchscopes is asked to confirm bridge-entity encoding rather than to gate head inclusion, and the heads under test have already survived a stronger causal filter.
A sensitivity analysis with stricter thresholds (Appendix~\ref{app:patchscopes-threshold}) shows that the Qwen $>$ Llama ordering on language-specific BRH counts is preserved at every threshold, and refines (rather than refutes) the cross-model dual-circuit picture: Llama distributes language-specific bridge encoding across many heads with intermittent activation, while Qwen concentrates it in a smaller set of more consistently active heads.

\noindent\textbf{Amplification range.}
Our amplification sweep stops at $\alpha = 1.0$.
The cross-lingual transfer curves on Spanish saturate near $\alpha = 0.5$, but the other languages still rise at $\alpha = 1.0$, so we may be reporting a lower bound on the achievable rescue rate.
Larger scales, or layer-conditioned scaling, may push these numbers further, though the cost to originally correct items also grows with $\alpha$.
\section*{Acknowledgments}
\label{s:ack}
This work was supported by the Institute of Information \& Communications Technology Planning \& Evaluation (IITP) grant funded by the Korean government (MSIT) (No. RS-2026-25530213, Development of Security Threat Analysis and Safety Validation Technologies for the LMM Ecosystem).

\nocite{*}
\bibliography{refs}
\appendix
\section{Data Structure}
\label{sec:data-structure}

\subsection{Dataset Construction}
\label{app:dataset:construct}

\noindent\textbf{Entity and relation coverage.}
The dataset extends \citet{biran2024hopping} to multilingual, with 17 semantic entity categories derived from Wikidata's \texttt{instance-of} (P31) hierarchy and 29 relation types.
Table~\ref{tab:relations} lists all 29 relations with per-hop occurrence counts; Table~\ref{tab:entity_types} summarises the entity-type distribution across the three chain positions ($A$, $B$, $C$).

\begin{table}[hbt!]
\centering
\small
\resizebox{\columnwidth}{!}{%
\begin{tabular}{llrr}
\toprule
\textbf{Property ID} & \textbf{Relation} & \textbf{$r_1$} & \textbf{$r_2$} \\
\midrule
P27   & country of citizenship    & 10,468 &  8,908 \\
P1441 & present in work           &  9,148 &  1,445 \\
P495  & country of origin         &  8,272 &  3,615 \\
P264  & record label              &  6,438 &  2,416 \\
P175  & performer                 &  6,057 &  1,109 \\
P108  & employer                  &  5,600 &    532 \\
P272  & production company        &  4,192 &  1,163 \\
P22   & father                    &  3,729 &  2,592 \\
P86   & composer                  &  3,573 &  1,151 \\
P178  & developer                 &  3,295 &    388 \\
P26   & spouse                    &  2,746 &  5,297 \\
P123  & publisher                 &  2,689 &    833 \\
P800  & notable work              &  2,596 &  3,235 \\
P50   & author                    &  2,204 &    940 \\
P127  & owned by                  &  2,068 &  4,059 \\
P112  & founded by                &  1,925 &  7,958 \\
P162  & producer                  &  1,503 &  1,351 \\
P58   & screenwriter              &  1,434 &  1,333 \\
P57   & director                  &  1,215 &    948 \\
P25   & mother                    &    832 &  2,101 \\
P159  & headquarters location     &    827 &  7,783 \\
P740  & location of formation     &    490 &  2,332 \\
P169  & chief executive officer   &    390 &  1,320 \\
P35   & head of state             &    267 &  7,862 \\
P36   & capital                   &     40 &  4,635 \\
P85   & anthem                    &     11 &  2,161 \\
P38   & currency                  &     -- &  4,141 \\
P1789 & chief operating officer   &      8 &     82 \\
P1304 & central bank              &      3 &    330 \\
\midrule
\multicolumn{2}{l}{\textit{Total instances}} & \multicolumn{2}{r}{82,020} \\
\bottomrule
\end{tabular}%
}
\caption{All 29 Wikidata relation types used in the dataset, with their property IDs and per-hop occurrence counts. A dash (--) indicates a relation that does not appear in that hop position.}
\label{tab:relations}
\end{table}

\begin{table*}[t]
\centering
\small
\begin{tabular}{lrrr}
\toprule
\textbf{Entity Type} & \textbf{Subject ($A$)} & \textbf{Bridge ($B$)} & \textbf{Answer ($C$)} \\
\midrule
Human / Person              & 26,906 & 23,804 & 33,963 \\
Country                     &    313 & 18,893 & 13,213 \\
Film                        & 12,176 &  7,309 &  1,415 \\
Film Character              &  8,699 &    760 &     72 \\
Video Game                  &  7,859 &  1,680 &  1,012 \\
Literary Character          &  7,699 &  1,458 &    360 \\
Musical Work/Composition    &  4,891 &    378 &  3,063 \\
Musical Group               &  3,286 &    500 &     29 \\
Literary Work               &  3,047 &  1,651 &  1,169 \\
Public Company              &  2,678 &  6,767 &  3,295 \\
Record Label                &  1,117 &  6,460 &  2,758 \\
Video Game Developer        &    913 &  5,530 &  1,442 \\
Novel Series                &    797 &    525 &    158 \\
Film Production Company     &    779 &  4,594 &  1,542 \\
City                        &    350 &  1,239 & 14,097 \\
Book Publisher              &    313 &    455 &    289 \\
Currency                    &     -- &     -- &  4,141 \\
Online Service              &    197 &     17 &      2 \\
\bottomrule
\end{tabular}
\caption{Entity-type distribution across the three chain positions. Counts indicate the number of instances in which an entity of that type occupies the given position.}
\label{tab:entity_types}
\end{table*}

\noindent\textbf{Query variants.}
Table~\ref{tab:query_types} provides concrete multilingual examples of all five query variants derived from each two-hop triple $(A, r_1, B, r_2, C)$.

\begin{table*}[t]
\centering
\small
\begin{tabular}{lll}
\toprule
\textbf{Query Type} & \textbf{Input} & \textbf{Example (English)} \\
\midrule
Two-Hop             & $A, r_1, r_2$   & \textit{What is the capital of the country of citizenship of Stephen Harper?} \\
First-Hop           & $A, r_1$        & \textit{What is the country of citizenship of Stephen Harper?} \\
Second-Hop          & $B, r_2$        & \textit{What is the capital of Canada?} \\
Shortcut (no $A$)   & $r_1, r_2$      & \textit{What is the capital of the country of citizenship of?} \\
Shortcut (no $r_1$) & $A, r_2$        & \textit{What is the capital of Stephen Harper?} \\
\bottomrule
\end{tabular}
\caption{Five query variants derived from each two-hop triple. The running example uses $A\!=\!\text{Stephen Harper}$, $r_1\!=\!\text{country of citizenship}$, $B\!=\!\text{Canada}$, $r_2\!=\!\text{capital}$, $C\!=\!\text{Ottawa}$.}
\label{tab:query_types}
\end{table*}

\subsection{Multilingual Extension}
\label{app:dataset:multilingual}

\noindent\textbf{Language-specific question templates.}
We pre-define relation templates for all 29 relations across all five languages, encoding language-specific elements such as nominal question words, genitive particles, postpositions, and copular verbs (Table~\ref{tab:multilingual_examples}).
The \texttt{[Answer]} readout token is kept as the ASCII string ``\texttt{Answer:}'' across all languages, ensuring a language-invariant readout position.
The instruction prefix is likewise localised per language (e.g., ``\textit{\foreignlanguage{korean}{정답만 짧게 답하시오.}}'' in Korean).

\begin{table*}[t]
\centering
\small
\setlength{\tabcolsep}{5pt}
\begin{tabular}{cl p{12.5cm}}
\toprule
\textbf{Lang.} & \textbf{Type} & \textbf{Example} \\
\midrule
\multirow{5}{*}{en}
  & Two-Hop     & \textit{What is the capital of the country of citizenship of Stephen Harper?} \\
  & First-Hop   & \textit{What is the country of citizenship of Stephen Harper?} \\
  & Second-Hop  & \textit{What is the capital of Canada?} \\
  & SC-no-$A$   & \textit{What is the capital of the country of citizenship of?} \\
  & SC-no-$r_1$ & \textit{What is the capital of Stephen Harper?} \\[4pt]
\multirow{5}{*}{ko}
  & Two-Hop     & \foreignlanguage{korean}{Stephen Harper의 국적의 수도는 무엇입니까?} \\
  & First-Hop   & \foreignlanguage{korean}{Stephen Harper의 국적은 무엇입니까?} \\
  & Second-Hop  & \foreignlanguage{korean}{Canada의 수도는 무엇입니까?} \\
  & SC-no-$A$   & \foreignlanguage{korean}{의 국적의 수도는 무엇입니까?} \\
  & SC-no-$r_1$ & \foreignlanguage{korean}{Stephen Harper의 수도는 무엇입니까?} \\[4pt]
\multirow{5}{*}{zh}
  & Two-Hop     & \foreignlanguage{chinese}{Stephen Harper的国籍的首都是什么？} \\
  & First-Hop   & \foreignlanguage{chinese}{Stephen Harper的国籍是什么？} \\
  & Second-Hop  & \foreignlanguage{chinese}{Canada的首都是什么？} \\
  & SC-no-$A$   & \foreignlanguage{chinese}{的国籍的首都是什么？} \\
  & SC-no-$r_1$ & \foreignlanguage{chinese}{Stephen Harper的首都是什么？} \\[4pt]
\multirow{5}{*}{ja}
  & Two-Hop     & \foreignlanguage{japanese}{Stephen Harperの国籍国の首都は何ですか？} \\
  & First-Hop   & \foreignlanguage{japanese}{Stephen Harperの国籍国は何ですか？} \\
  & Second-Hop  & \foreignlanguage{japanese}{Canadaの首都は何ですか？} \\
  & SC-no-$A$   & \foreignlanguage{japanese}{の国籍国の首都は何ですか？} \\
  & SC-no-$r_1$ & \foreignlanguage{japanese}{Stephen Harperの首都は何ですか？} \\[4pt]
\multirow{5}{*}{es}
  & Two-Hop     & \textit{¿Cuál es la capital del país de ciudadanía de Stephen Harper?} \\
  & First-Hop   & \textit{¿Cuál es el país de ciudadanía de Stephen Harper?} \\
  & Second-Hop  & \textit{¿Cuál es la capital de Canada?} \\
  & SC-no-$A$   & \textit{¿Cuál es la capital del país de ciudadanía de?} \\
  & SC-no-$r_1$ & \textit{¿Cuál es la capital de Stephen Harper?} \\
\bottomrule
\end{tabular}
\caption{One representative chain (\mbox{Stephen Harper $\xrightarrow{\text{country}}$ Canada $\xrightarrow{\text{capital}}$ Ottawa}) shown in all five query variants for each language. Entity labels are retrieved from Wikidata; interrogative framing is generated via per-language relation templates. \textit{SC-no-$A$}: shortcut with subject removed; \textit{SC-no-$r_1$}: shortcut with first-hop relation removed.}
\label{tab:multilingual_examples}
\end{table*}

\section{Layer-wise Patching: Logit-Difference Results}
\label{sec:layer-wise-patching}

\begin{figure}[t]
  \centering
  \includegraphics[width=\linewidth]{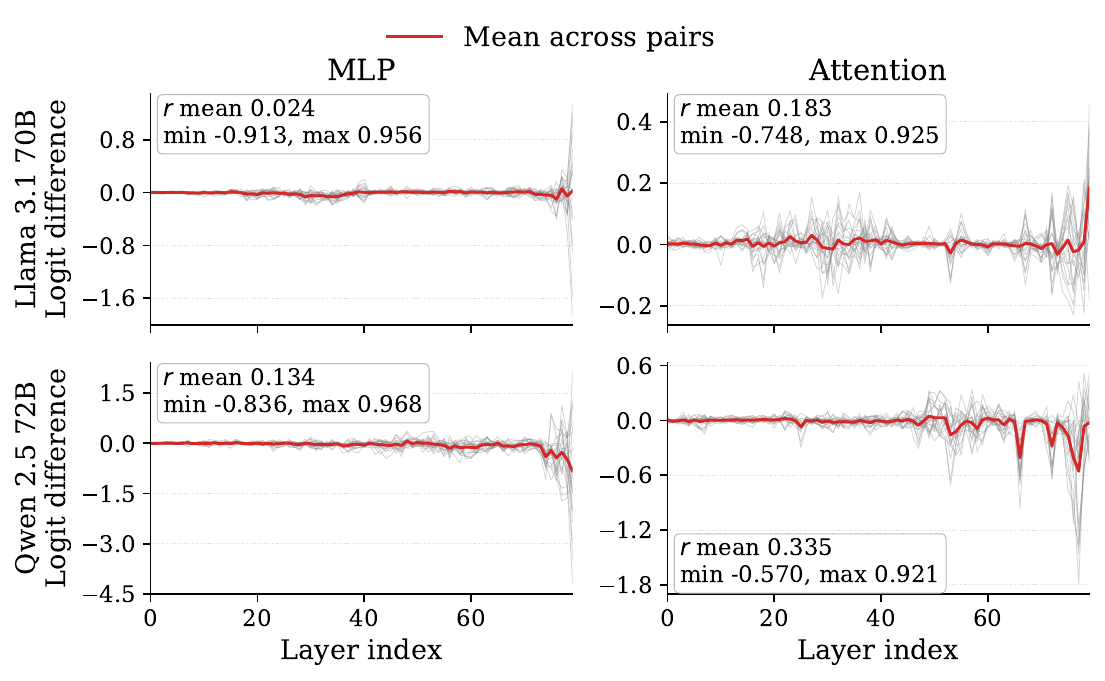}
  \caption{Layer-wise activation patching, logit difference at \texttt{answer\_colon}.
  Layout mirrors Figure~\ref{fig:stage1-sweep}; each panel overlays all 20 directional language-pair curves (grey) and their mean (red).
  The inset reports the mean and range of off-diagonal Pearson $r$ across the 380 language-pair cross-correlations.}
  \label{fig:stage1-sweep-ld}
\end{figure}

Figure~\ref{fig:stage1-sweep-ld} shows the per-layer LD curves for the same activation-patching experiment reported in Section~\ref{s:exp-stage1}, complementing the KL-divergence results in the main text.

\noindent\textbf{MLP LD is near-zero and cross-pair correlations collapse.}
In striking contrast to the KL profiles---where MLP off-diagonal correlations reach $r = 0.989$ (Llama) and $0.936$ (Qwen)---the LD correlations across MLP patches are near chance: $r = 0.02$ (range $-0.91$--$0.96$) for Llama and $0.13$ ($-0.84$--$0.97$) for Qwen.
The mean LD curve stays close to zero for virtually all intermediate layers and only dips sharply negative at the very last layers, most prominently in Qwen ($\approx -3.5$ at the final layer).
This dissociation between the two metrics is informative: MLP patches produce \emph{consistently large distribution changes} (high KL) but their effect on the clean answer token is \emph{highly variable and pair-dependent} (low LD correlation).
The negative late-layer LD is consistent with a source-language MLP activation being injected into a target-language context—it shifts the output distribution substantially (hence high KL) while actively suppressing the target-language answer logit, since the stored factual representation is aligned to the source language's output direction.

\noindent\textbf{Attention LD is heterogeneous, mirroring the KL fan-out.}
Attention LD correlations are likewise low ($r = 0.18$, range $-0.75$--$0.93$ in Llama; $r = 0.34$, $-0.57$--$0.92$ in Qwen), but the individual curves exhibit visibly wider spread across the mid-to-late layer range, echoing the KL fan-out that motivated the head-level analysis in Stage 2.
Qwen shows markedly larger suppressive effects ($\approx -1.8$) concentrated around layers 60--70, whereas Llama attention remains near zero with modest fluctuations ($\approx \pm 0.4$) throughout.
The higher absolute magnitude in Qwen suggests that its attention mechanism exerts a stronger gating influence on the output token, and that this gating is particularly sensitive to cross-lingual context at mid-to-late layers.

\section{BRS Contrast Sensitivity}
\label{app:brs-contrast}

We test whether the BRS contrast $\mathrm{BRS}(h) = z(\mathrm{FH}) + z(\mathrm{TH}) - z(\mathrm{SH})$ drives the pipeline output by comparing it to three alternative contrasts on the same gradient measurements: \texttt{TH} (two-hop term alone), \texttt{TH$-$SH} (drop the FH term), and \texttt{FH+TH$-$2SH} (penalize SH more heavily).
For each variant we recompute per-head scores, form per-language pools at the same top-$N$\% threshold, and intersect across the five languages to obtain a general candidate pool.
Table~\ref{tab:brs-contrast} reports per-head Spearman rank correlation against the original BRS (mean over the 10 \{general, language-specific$\times$4\} sets per model), the general pool sizes, and the recall of the original general pool inside each variant pool.

\begin{table}[t]
  \centering
  \scriptsize
  \setlength{\tabcolsep}{4pt}
  \begin{tabular}{l c c c}
    \hline
    Variant & \shortstack{Spearman\\$\rho$} & \shortstack{Pool size\\(L / Q)} & Recall \\
    \hline
    $\mathrm{FH{+}TH{-}SH}^\dagger$ & 1.000 & 191 / 175 & --- \\
    $\mathrm{TH}$ & \textbf{0.962} & 305 / 291 & \textbf{84--89\%} \\
    $\mathrm{TH{-}SH}$ & 0.328 & 149 / 98  & 26--35\% \\
    $\mathrm{FH{+}TH{-}2SH}$ & 0.388 & 102 / 85  & 27--28\% \\
    \hline
  \end{tabular}
  \caption{Sensitivity of the BRS pipeline to alternative contrasts. $^\dagger$ denotes the original contrast used in the main paper. Spearman $\rho$ is the mean per-head rank correlation against the original BRS over the 10 \{general, KO/ZH/JA/ES-specific\}$\times$2-model sets. Pool size reports the Llama and Qwen general candidate pool sizes (L\,/\,Q). Recall is the fraction of the original general pool retained inside each variant's general pool, given as the Llama--Qwen range.}
  \label{tab:brs-contrast}
\end{table}

Two observations follow.
(i) The TH term alone produces near-identical per-head rankings (mean Spearman $\rho = 0.962$), indicating that the two-hop task gradient is the dominant signal in the BRS contrast.
(ii) Despite this, the TH-only candidate pool is 60\% larger than the original (305 vs.\ 191 heads in Llama general; 291 vs.\ 175 in Qwen) yet still contains 84--89\% of the original general pool as a subset.
The FH$+$ and $-$SH terms therefore do not change \emph{which} heads are most important; they narrow the candidate pool to the \emph{chaining-specific} subset by requiring heads to also matter for first-hop inference (FH$+$) while \emph{not} mattering for the externally supplied second-hop ($-$SH).
This separation---importance ranking from chaining-specificity---matches the design rationale stated in \S\ref{s:stage2}.
The alternative \texttt{TH$-$SH} and \texttt{FH+TH$-$2SH} contrasts reorder heads (Spearman $\rho < 0.4$) and retain only 26--35\% of the original general pool, confirming that the specific weighting of the FH and SH terms shapes the chaining-specific selection in a non-trivial way.

\section{Source-Entity Localization Control}
\label{app:entity-localization}

The main pipeline keeps the source entity $A$ in its English surface form across all five languages (\S\ref{s:data}), so every non-English prompt carries an English entity string.
We test whether the language-general pool depends on that choice, since a pool defined by an English token shared across all languages could reflect a common response to the token rather than shared chaining machinery.
We localize $A$ into its target-language surface form, recompute the per-head BRS on both models, and re-form the per-language pools at the same top-$N$\% threshold before intersecting across the five languages, holding every other component of the pipeline fixed.

Two observations follow.
(i) Per-head rankings are essentially unchanged: the Spearman rank correlation against the original BRS over all $5{,}120$ heads is $\rho = 0.89$--$0.98$ for Llama and $0.92$--$0.97$ for Qwen across KO/ZH/JA/ES.
(ii) The localized run re-identifies most of the original general pool---134 of 191 heads (70\%) in Llama and 121 of 175 (69\%) in Qwen---and the residual difference is expected under a top-$N$\% intersection, where heads near the threshold boundary move in or out under any perturbation.
The language-general routing pattern is therefore not an artifact of keeping entity names in English.

\section{Amplification Cost}
\label{app:amplification-cost}

The main text evaluates amplification only on items the model originally answered incorrectly (\S\ref{s:stage3}), which leaves its effect on already-correct predictions unmeasured.
We therefore re-run the same intervention on two-hop item sets that include already-correct items (Table~\ref{tab:amp-cost}) and on English MMLU (Table~\ref{tab:amp-mmlu}), the latter for every amplified head set rather than the general pool alone.

\begin{table}[t]
  \centering
  \footnotesize
  \setlength{\tabcolsep}{3.5pt}
  \begin{tabular}{l l c c c c c c}
    \hline
    & & \multicolumn{3}{c}{Correct-only} & \multicolumn{3}{c}{Balanced} \\
    Model & Lang & .25 & .5 & 1.0 & .25 & .5 & 1.0 \\
    \hline
    Llama & ko & 95 & 92 & 90 & 65 & 65 & 71 \\
          & zh & 96 & 92 & 93 & 60 & 68 & 75 \\
          & ja & 96 & 90 & 87 & 55 & 62 & 65 \\
          & es & 100 & 94 & 88 & 66 & 63 & 57 \\
          & \textsc{all} & 96.8 & 92.0 & 89.5 & 61.5 & 64.5 & 67.0 \\
    \hline
    Qwen  & ko & 96 & 89 & 83 & 58 & 59 & 57 \\
          & zh & 96 & 92 & 91 & 55 & 56 & 61 \\
          & ja & 97 & 93 & 92 & 51 & 51 & 54 \\
          & es & 93 & 88 & 83 & 60 & 61 & 61 \\
          & \textsc{all} & 95.5 & 90.5 & 87.3 & 56.0 & 56.8 & 58.3 \\
    \hline
  \end{tabular}
  \caption{Two-hop accuracy (\%) under general-BRH amplification at $\alpha \in \{0.25, 0.5, 1.0\}$ (column headers). \emph{Correct-only}: 100 items per language the unmodified model answers correctly, so the $\alpha = 0$ baseline is 100\% by construction. \emph{Balanced}: 100 items per language, 50 originally correct and 50 originally incorrect, so the baseline is 50\%. \textsc{all} pools the four languages (400 items).}
  \label{tab:amp-cost}
\end{table}

\begin{table}[t]
  \centering
  \footnotesize
  \setlength{\tabcolsep}{5pt}
  \begin{tabular}{l l c c c c}
    \hline
    Model & Set & $k$ & .25 & .5 & 1.0 \\
    \hline
    Llama & general & 70 & $-0.5$ & $-1.6$ & $-1.6$ \\
          & ko-spec &  8 & $-0.5$ & $+0.5$ & $-0.5$ \\
          & zh-spec & 11 & $-1.0$ & $-0.5$ & $-0.5$ \\
          & ja-spec & 31 & $-2.1$ & $-1.6$ & $-2.1$ \\
          & es-spec &  9 & $0.0$ & $0.0$ & $0.0$ \\
    \hline
    Qwen  & general & 42 & $0.0$ & $-0.5$ & $-0.5$ \\
          & ko-spec & 87 & $-0.5$ & $-0.5$ & $+0.5$ \\
          & zh-spec & 38 & $+0.5$ & $0.0$ & $-0.5$ \\
          & ja-spec & 28 & $0.0$ & $0.0$ & $0.0$ \\
          & es-spec & 31 & $+0.5$ & $-0.5$ & $-0.5$ \\
    \hline
  \end{tabular}
  \caption{Change in English MMLU accuracy (percentage points) with amplification active, by amplified head set and $\alpha$ (column headers). MMLU is 5-shot over 194 items; the unmodified baselines are 81.96\% (Llama) and 88.66\% (Qwen). $k$ is the number of heads in the set.}
  \label{tab:amp-mmlu}
\end{table}

Three observations follow.
(i) \emph{Amplification costs correct items, and the cost grows with $\alpha$.}
On the correct-only set, accuracy declines monotonically in $\alpha$, reaching 89.5\% (Llama) and 87.3\% (Qwen) at $\alpha = 1.0$, with a per-language spread of 87--93\% and 83--92\% respectively.
(ii) \emph{On a mixed pool the gains outweigh the losses at every $\alpha$.}
At $\alpha = 1.0$ on the balanced set, amplification recovers 90 items and loses 22 in Llama, and recovers 57 and loses 24 in Qwen; the net gain rises with $\alpha$ in both models even as the correct-only cost grows.
Because that pool's composition is fixed by construction rather than reflecting the model's actual error rate, these figures characterise the trade-off the intervention makes, not an expected gain under deployment, where the failing items are not known in advance.
(iii) \emph{General-domain ability is largely unaffected.}
English MMLU changes by at most 2.1 points across all ten (model, head set) pairs, and by $-1.6$ (Llama) and $-0.5$ (Qwen) points for the general pool at $\alpha = 1.0$; the largest single drop, $-2.1$ points, occurs for Llama's Japanese-specific set rather than for either general pool.

\section{Patchscopes Threshold Sensitivity}
\label{app:patchscopes-threshold}

The main pipeline applies a permissive criterion (\S\ref{s:stage2}): a head passes if the gold bridge $B$ appears in at least one trial across the Patchscopes prompts (denoted $k=0$ below).
Table~\ref{tab:patchscopes-threshold} reports how the final language-specific BRH counts change under progressively stricter success-count thresholds: $k=1$ requires success in $\geq 3$ trials, $k=2$ in $\geq 4$, and $k=3$ in $\geq 5$.

\begin{table}[t]
  \centering
  \footnotesize
  \setlength{\tabcolsep}{5pt}
  \begin{tabular}{l l c c c c}
    \hline
    Model & Lang & \shortstack{$k=0$\\(main)} & $k=1$ & $k=2$ & $k=3$ \\
    \hline
    Llama & ko &  8 &  1 &  1 & 0 \\
          & zh & 11 &  0 &  0 & 0 \\
          & ja & 31 & 13 & 13 & 1 \\
          & es &  9 &  0 &  0 & 0 \\
    \hline
    Qwen  & ko & 87 & 49 & 49 & 10 \\
          & zh & 38 &  8 &  8 & 0 \\
          & ja & 28 &  4 &  4 & 0 \\
          & es & 31 & 10 & 10 & 0 \\
    \hline
  \end{tabular}
  \caption{Final language-specific BRH counts under progressively stricter Patchscopes thresholds. $k$ indexes the success-count requirement: $k=0$ is the main paper default ($\geq 1$ successful trial); $k=1$, $2$, $3$ require $\geq 3$, $\geq 4$, $\geq 5$ successful trials respectively. $k=1$ and $k=2$ yield identical counts because no candidate head has exactly $3$ successful trials.}
  \label{tab:patchscopes-threshold}
\end{table}

Two observations follow.
(i) \emph{The Qwen $>$ Llama ordering is preserved at every threshold.} Qwen retains substantially more language-specific BRH than Llama at every $k$; the Korean ratio grows from $87/8 \approx 11{\times}$ (at $k=0$) to $49/1 \approx 49{\times}$ (at $k=1$) and $10/0$ (at $k=3$, undefined). The cross-model dual-circuit dichotomy reported in \S\ref{s:exp-stage3} is therefore robust to the Patchscopes criterion.
(ii) \emph{The two models encode language-specific bridge information differently.} Llama's language-specific BRH largely disappear at $k=1$ (only Japanese retains a non-trivial 13 heads), suggesting that Llama distributes bridge encoding across many heads with intermittent, probabilistic activation. Qwen's pools degrade more gracefully---especially Korean, where 49 heads remain at $k=1$ and 10 at $k=3$---suggesting that Qwen concentrates bridge encoding in a smaller set of consistently active heads.
This refines (rather than refutes) the dual-circuit picture: Llama's reliance on the general pool is paired with diffuse, low-redundancy language-specific encoding, whereas Qwen's larger language-specific pools are also individually more robust encoders of the bridge entity.

\section{Patchscopes Example Outputs}
\label{sec:appendix-patchscopes}

Table~\ref{tab:patchscopes} lists representative Patchscopes outputs for language-specific BRH across both models and all four non-pivot languages.
The patched hidden state is injected at the \texttt{x:} position of a few-shot interpretation prompt, and the gold bridge entity $B$ appears in every continuation.
Surface tokens from the in-context examples can leak into the output as a benign in-context artifact, which we ignore when counting a head as successful.

\section{Head-Level BRS Landscape}
\label{sec:bridge-routing-heads-heatmap}

\begin{figure*}[!t]
  \centering
  \includegraphics[width=\linewidth]{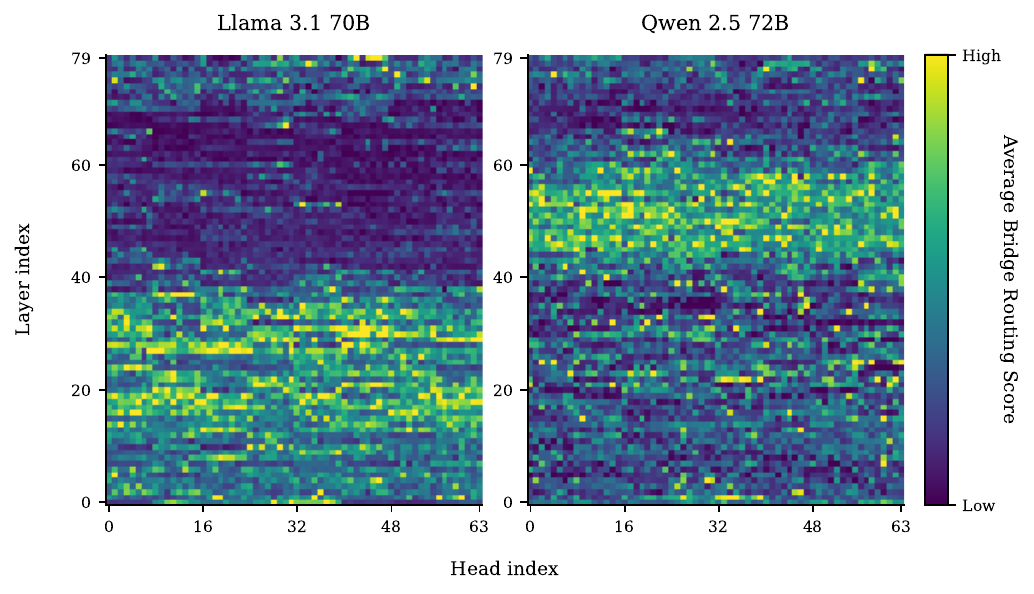}
  \caption{Bridge Routing Score (BRS) heatmaps for Llama 3.1 70B (left) and Qwen 2.5 72B (right).
    Color encodes the per-head BRS averaged across all five languages, computed before any filtering step (yellow = high, dark purple = low).
    The $x$-axis is the head index (0--63) and the $y$-axis is the layer index (0--79), with layer~0 at the bottom.}
  \label{fig:brh-heatmap}
\end{figure*}

Figure~\ref{fig:brh-heatmap} shows the unthresholded BRS landscape across the full 80-layer $\times$ 64-head grid of each model, providing a pre-filtering diagnostic that complements the layer-level Stage~1 results.

\noindent\textbf{The BRS landscape as a pre-filtering diagnostic.}
The heatmap renders the unthresholded BRS signal as a continuous importance map; LAHIS~\cite{Liu_Song_Zhou_Du_Xu_Jiang_Zhang_Jia_2026} estimates per-head importance as $\mathcal{I}_h = |\partial\mathcal{L}/\partial m_h|$ at the unit mask $m_h = 1$, yielding a first-order approximation to the counterfactual effect of suppressing each head at the cost of only one forward--backward pass per condition---$O(1)$ overhead against the $O(H)$ cost of exhaustive ablation over all $H = 5{,}120$ heads.
The per-condition z-normalization removes systematic magnitude differences across conditions, ensuring the TH$+$FH$-$SH contrast reflects \emph{relative} routing specificity rather than raw gradient scale.

\noindent\textbf{Layer concentration mirrors Stage 1 attention divergence.}
The BRS heatmap reveals that high-importance heads are not uniformly distributed but form elevated horizontal bands concentrated in the middle-to-upper layer range---roughly layers 15--45 for Llama and layers 40--75 for Qwen.
This spatial pattern closely tracks the mid-to-late layer range in which attention KL profiles begin to fan out across language pairs in Figure~\ref{fig:stage1-sweep}.
The agreement is notable because the two analyses use entirely different signals: Stage 1 measures residual-stream perturbation at the layer level, whereas the BRS heatmap measures head-level gradient importance under a differential condition design.
The heatmap therefore provides a granular, pre-filtering complement to the layer-aggregated Stage 1 result: both independently localize cross-lingual bridge routing to the same portion of the network.

\noindent\textbf{BRH placement within the landscape.}
Comparing this BRS landscape with the BRH layer distribution in Figure~\ref{fig:brh-layers} shows that the final heads consistently fall within the elevated-BRS bands, confirming that the three-stage filter selects heads genuinely prominent in the unthresholded landscape rather than incidentally surviving a threshold.

\begin{table*}[t]
  \centering
  \footnotesize
  \setlength{\tabcolsep}{4pt}
  \begin{tabular}{l l p{5.0cm} l p{4.8cm}}
    \hline
    Model & Lang & Question (target language / English gloss) & Gold $B$ & Generated continuation \\
    \hline
    Llama 3.1 70B & ko & \foreignlanguage{korean}{Rod Laver의 국적의 수도는 무엇입니까?} \par {\scriptsize\itshape What is the capital of Rod Laver's nationality?} & Australia & ``Australia: Australia is a country in the southern hemisphere, ...'' \\
                  & zh & \foreignlanguage{chinese}{Mark Thatcher的母亲的配偶是谁？} \par {\scriptsize\itshape Who is the spouse of Mark Thatcher's mother?} & M.\ Thatcher & ``Margaret Thatcher: Margaret Thatcher was the Prime Minister of the UK ...'' \\
                  & ja & \foreignlanguage{japanese}{Psychoのプロデューサーの配偶者は誰ですか？} \par {\scriptsize\itshape Who is the spouse of Psycho's producer?} & A.\ Hitchcock & ``Hitchcock: Hitchcock is a film directed by Alfred Hitchcock ...'' \\
                  & es & ¿Cuál es la capital del país de ciudadanía de Alexander Arekeev? \par {\scriptsize\itshape What is the capital of Alexander Arekeev's country of citizenship?} & Russia & ``Russia: Russia is a country located in northern Eurasia ...'' \\
    \hline
    Qwen 2.5 72B  & ko & \foreignlanguage{korean}{James Sirius Potter의 등장 작품의 출판사는 누구입니까?} \par {\scriptsize\itshape Who is the publisher of the work in which James Sirius Potter appears?} & Harry Potter & ``Harry Potter is the protagonist of the Harry Potter series written by J.K. Rowling ...'' \\
                  & zh & \foreignlanguage{chinese}{Elsa Einstein的配偶的配偶是谁？} \par {\scriptsize\itshape Who is the spouse of Elsa Einstein's spouse?} & A.\ Einstein & ``spouse, Albert Einstein: Albert Einstein was a German-born theoretical physicist ...'' \\
                  & ja & \foreignlanguage{japanese}{Windows Updateの開発者の創設者は誰ですか？} \par {\scriptsize\itshape Who is the founder of the developer of Windows Update?} & Microsoft & ``Samsung manufactures a computer that uses Windows XP Pro. Windows XP Pro was created by Microsoft Corporation ...'' \\
                  & es & ¿Cuál es la moneda del país de ciudadanía de Microchip? \par {\scriptsize\itshape What is the currency of Microchip's country of citizenship?} & USA & ``of $x$, United States of America: The United States of America (USA), commonly known as ...'' \\
    \hline
  \end{tabular}
  \caption{Patchscopes examples for language-specific BRH. Each row shows the target-language question (top) with an English gloss (bottom, italics). The gold bridge entity $B$ appears explicitly in every continuation.}
  \label{tab:patchscopes}
\end{table*}

\end{document}